\documentclass[letterpaper,10pt,conference]{ieeeconf}
\IEEEoverridecommandlockouts
\usepackage{amsmath,amssymb}
\usepackage{mathtools}
\usepackage{bm}
\usepackage{subfig}
\usepackage{xcolor}
\usepackage{graphicx}
\usepackage{algorithm}
\usepackage{algpseudocode}
\usepackage{booktabs}

\newcommand{\norm}[1]{\left\|#1\right\|}

\title{\LARGE \bf
	Density-Driven Area Coverage for Nonholonomic Multi-Robot Systems with Safety Guarantee}

\author{Julian Martinez$^1$ and Kooktae Lee$^1$
	\thanks{*This work was supported by NSF CAREER Grant CMMI-DCSD-2638508.}
    \thanks{$^{1}$Julian Martinez and Kooktae Lee are with the Department of Mechanical and Aerospace Engineering, Texas Tech University, Lubbock, TX 79409, USA, email: mar85781@ttu.edu, kooktae.lee@ttu.edu} 
}

\begin{document}
	\maketitle
	
	\begin{abstract}
Density-Driven Optimal Control (D$^2$OC) provides a principled approach to
distributing multi-robot teams over non-uniform spatial distributions.
Applying D$^2$OC to nonholonomic robots, however, creates a gap between
safety constraints imposed on a reference motion and the physical inputs
that determine the actual robot motion. We address this issue by enforcing
the safety constraint directly on the robot's physical inputs while
preserving the density-driven coverage objective. The proposed framework
combines D$^2$OC with a control barrier function safety filter through a
feedback-linearizing look-ahead point, allowing safety and actuator limits
to be considered together during control. We further derive a safety
margin that accounts for the look-ahead geometry, robot footprint, and
motion during each control interval. Simulation results show that the
proposed method maintains the required physical separation while achieving
coverage performance comparable to a conventional reference-tracking
approach, which can satisfy safety on the reference motion yet violate the
corresponding physical clearance. Experiments on multiple nonholonomic
robots in the Robotarium further demonstrate safe execution while driving
the robots toward the desired spatial distribution. These results show
that enforcing safety directly on the physical inputs can eliminate the
mismatch between safety certification and physical robot motion in
density-driven multi-robot coverage.
\end{abstract}

\section{Introduction}
Multi-robot coverage of non-uniform spatial distributions is important in
environmental monitoring, precision agriculture, and autonomous inspection, which is
classically addressed via Voronoi-based coverage control
\cite{cortes2004coverage}. Recent work on PDE-constrained density control formulations
offers alternative frameworks for distribution shaping \cite{niu2025decentralized}.
In parallel, Density-Driven Optimal Control (D$^2$OC) \cite{seo2025d2oc,seo2026farm}
formulates this problem as optimal-transport-based distribution matching,
driving agents toward a prescribed density while accounting for their dynamics. Further,
its extensions have considered connectivity, stochastic dynamics, and
application-specific coverage requirements
\cite{lee2026tac,lee2026stochastic,lee2026ai}. Applying such coverage
methods to planar nonholonomic robots, however, introduces a control
mismatch. The planar motion generated by the coverage law cannot be directly
commanded. In this work, we consider the standard unicycle model as the
underlying nonholonomic robot, whose control input is $(v,\omega)$, with
forward and angular velocities but no direct lateral velocity command.

A common implementation is therefore to generate a planar reference and use
a separate controller to track it
\cite{de2000stabilization}. While this separation simplifies motion
generation, it also separates the desired motion from the physical inputs
that determine the actual robot motion. This distinction becomes critical
when safety and actuator limits are considered. A safety constraint imposed
on the reference motion does not directly constrain the physical actuation,
and the available control authority can limit the ability to realize a
required corrective motion.

To overcome these challenges, Control Barrier Functions (CBFs) can be
adopted to provide a natural framework for incorporating safety constraints
into feedback control \cite{ames2019cbf,jankovic2018unicycle}. CBF-based
quadratic programs have been widely applied to multi-agent collision
avoidance and safety-critical control
\cite{wang2017cbf-multi,notomista2019cvt}. For input-constrained systems,
however, safety depends not only on the barrier condition but also on
whether a feasible input exists within the actuator bounds
\cite{zhang2024feasibility,rabiee2025soft}. This feasibility challenge is
particularly relevant for nonholonomic UGVs, where the available steering
and thrust inputs can limit the ability of conventional CBF formulations
to generate collision-avoiding motions \cite{goswami2024collision}.
This motivates enforcing the safety constraint at the physical-input level
rather than on an intermediate reference trajectory.

We address this issue by expressing the D$^2$OC motion and the CBF
constraint at a feedback-linearizing look-ahead point. The velocity of this
point is invertibly related to the physical inputs $(v,\omega)$
\cite{lavalle2006planning}, allowing the CBF-QP to modify the nominal
coverage motion directly in the physical control space. Actuator limits are
therefore included in the same optimization that enforces safety, without
requiring a separate safety-critical tracking layer.

We therefore propose a \emph{single-layer physical-input safety framework}
for D$^2$OC coverage of nonholonomic robots, using the unicycle model as the
underlying robot model. The key idea is to formulate the coverage motion and
safety constraint at a common feedback-linearizing point, which allows
the CBF-QP to directly map the nominal coverage command to admissible
physical inputs while enforcing safety and actuator constraints in the same
optimization.

Thus, our main contributions
are:
\begin{itemize}
    \item a single-layer integration of D$^2$OC, finite-horizon LQ tracking,
    feedback linearization, and a physical-input CBF-QP for unicycle robots;
    \item a closed-form safety margin that accounts for the look-ahead offset,
    robot footprint, and motion during each control interval;
\item a two-layer reference-tracking baseline, using the same D$^2$OC
allocation and nominal controller, for direct comparison with the proposed
single-layer architecture and evaluation of the effect of enforcing safety
only on the virtual trajectory; and

    \item hardware validation on five nonholonomic robots in the Robotarium
    \cite{pickem2017robotarium}, using the platform's  actuator limits
    to evaluate the proposed safety margin.
\end{itemize}

\section{Problem Formulation}

We consider a team of $n_a$ nonholonomic unicycle robots tasked with
matching the time-averaged spatial distribution of their trajectories to a
prescribed target density using the D$^2$OC framework. While D$^2$OC supports
general linear time-invariant (LTI) agent dynamics, realizing its allocation
output on a nonholonomic platform requires respecting the underlying
kinematic constraints and physical inputs. We therefore retain the original
allocation mechanism and develop a control realization for unicycle robots
based on a feedback-linearizing look-ahead point, with safety constraints
formulated directly in the physical control space.

Throughout this paper, the robot dynamics and safety conditions are
formulated in continuous time, whereas the D$^2$OC allocation and nominal
tracking controller are updated at discrete sampling instants. We use $t$ for
continuous-time quantities and $t_k=kT$ (or equivalently the index $k$) for
sampled quantities. The two descriptions are connected through the
sampled-data implementation, in which the nominal command and safety filter
are evaluated at each sampling instant and the resulting physical input is
held over the corresponding sampling interval.

To make the nonholonomic realization explicit, we model robot $i\in\{1,2,\ldots,n_a\}$, where $n_a$ is the total number of robots, as a
unicycle with position $r_i=[r_{i,x},r_{i,y}]^\top$ and heading $\theta_i$,
whose dynamics are
\begin{equation}
\dot r_i =
v_i
\begin{bmatrix}
\cos\theta_i,
\sin\theta_i
\end{bmatrix}^{\top},
\qquad
\dot\theta_i=\omega_i,
\label{eq:unicycle}
\end{equation}
with bounded physical inputs
\begin{equation}
(v_i,\omega_i)\in
\mathcal U
=
[v_{\min},v_{\max}]
\times
[-\omega_{\max},\omega_{\max}].
\label{eq:input_constraint}
\end{equation}
Unlike a holonomic point, the robot cannot independently choose the
direction of its instantaneous translational velocity. It can only move
along its current heading. Thus, the realization of a desired planar motion
is inherently coupled to the physical inputs $(v_i,\omega_i)$.

In our experiments, we impose $v_{\min}=0$, so that the robots cannot
reverse. This restriction limits the available translational authority for
collision avoidance and makes the angular-rate bound $\omega_{\max}$
particularly important for generating feasible evasive maneuvers. This
constraint will be explicitly reflected in the safety margin and feasibility
analysis developed later.
	
	\subsection{Look-Ahead Feedback Linearization}

The rear-axle position $r_i$ cannot be assigned an arbitrary planar
velocity through the physical inputs $(v_i,\omega_i)$. In particular,
its instantaneous velocity is constrained to the robot's current heading,
while the angular velocity $\omega_i$ changes the heading rather than
directly providing a lateral velocity component. Consequently, the
planar motion of $r_i$ cannot be specified independently in the two
Cartesian directions. This creates a mismatch between the motion variables
used by the D$^2$OC tracking layer and the admissible motion of the
unicycle.

To obtain a directly actuated planar motion variable, we introduce a
look-ahead point
\begin{equation}
	p_i=r_i+d
	\begin{bmatrix}
		\cos\theta_i\\
		\sin\theta_i
	\end{bmatrix},
	\qquad d>0,
	\label{eq:lookahead}
\end{equation}
whose velocity satisfies the \textit{first-order feedback-linearization} relation
\begin{equation}
	\dot p_i
	=v_i
	\begin{bmatrix}
		\cos\theta_i\\
		\sin\theta_i
	\end{bmatrix}
	+d\omega_i
	\begin{bmatrix}
		-\sin\theta_i\\
		\cos\theta_i
	\end{bmatrix}
	=G(\theta_i)u_i,
	\;
	u_i=\begin{bmatrix}v_i\\\omega_i\end{bmatrix},
	\label{eq:fbl}
\end{equation}
where
\begin{equation}
	G(\theta_i)=
	\begin{bmatrix}
		\cos\theta_i&-d\sin\theta_i\\
		\sin\theta_i&d\cos\theta_i
	\end{bmatrix}.
	\label{eq:A}
\end{equation}
Since $\det G(\theta_i)=d\neq0$, the mapping between $\dot p_i$ and the
physical input $u_i$ is invertible for every heading. Thus, the look-ahead
point provides a first-order planar interface through which the desired
motion generated by the D$^2$OC-based controller can be realized directly
by the physical inputs. Importantly, this transformation does not modify
the D$^2$OC allocation rule; it provides the interface between the allocated
motion and the nonholonomic robot dynamics. All subsequent tracking and
safety constraints are therefore formulated using $p_i$, while the
geometric relation between $p_i$ and the physical robot position $r_i$ is
accounted for in the safety margin.

\subsection{D$^2$OC Allocation}

D$^2$OC provides a decentralized way to distribute multiple robots according
to a desired spatial density \cite{seo2025d2oc,seo2026farm,lee2026tac,
lee2026stochastic,lee2026ai}. Intuitively, the target density, represented by 
spatial samples $\{q_j\}_{j=1}^{N_{\rm sp}}$, specifies where robots should 
spend more or less time: regions with higher density should receive more 
coverage, while regions with lower density should receive less. 
Initially, each sample point $q_j$ is assigned a uniform weight inversely proportional 
to the total number of samples $N_{\rm sp}$ (i.e., $\frac{1}{N_{\rm sp}}$).
Instead of dividing the workspace into fixed regions, D$^2$OC
continuously determines which part of the target density each robot should
cover based on its current location and the remaining unallocated density.
As agents perform area coverage to match the target distribution, these sample 
weights are progressively depleted within each robot's local neighborhood $\mathcal{S}_i$. 
At each update, robot $i$ is assigned a portion of this remaining target density, 
summarized by the spatial reference
\begin{equation}
    q_i^\star=
    \frac{\sum_{j\in\mathcal S_i}\pi_jq_j}
    {\sum_{j\in\mathcal S_i}\pi_j},
    \qquad
    \gamma_i=\sum_{j\in\mathcal S_i}\pi_j ,
    \label{eq:d2oc-target}
\end{equation}
where $q_j$ denotes a spatial sample and $\pi_j$
denotes the transport plan allocating mass from sample $j$ to the agent. Thus, $q_i^\star$
is the weighted center of the mass currently assigned to robot $i$,
while $\gamma_i$ represents the total mass allocated by the transport plan at that update. 
As the robots move, these assignments are dynamically updated so that the robots
progressively cover the target density over time.

The important point for this work is that D$^2$OC produces a desired spatial
motion, while the underlying robot still has to physically realize that
motion. We therefore apply the allocation using the look-ahead point $p_i$
introduced before. The resulting D$^2$OC reference is consequently expressed
in the same spatial coordinate used by the tracking and safety controllers,
providing a direct interface between density-based coverage and the
nonholonomic robot dynamics.
	
\section{Controller Synthesis via Feedback Linearization}

The allocation layer produces, at every step, a target point $q_i^\star$
and a confidence weight $\gamma_i$ for each agent. However, it does not account 
for the unicycle's actuator limits or collision avoidance. To bridge this gap 
within a unified single-layer framework, we synthesize a control pipeline 
that directly maps allocations to physical actuators. First, a finite-horizon 
linear-quadratic (LQ) regulator treats the look-ahead point $p_i$ as a double 
integrator to generate a nominal Cartesian acceleration driving $p_i$ toward $q_i^\star$. 
Second, this acceleration is translated into physical input rates via exact feedback 
linearization and projected directly onto the physical actuator box $\mathcal{U}$ and 
collision-avoidance constraints via a Control Barrier Function Quadratic Program (CBF-QP). 
Executing the safety filter directly in the physical input space, rather than on an 
intermediate virtual trajectory, ensures that actuation limits and safety bounds are 
resolved within a single optimization stage.
	
	\subsection{Finite-Horizon LQ Nominal Control}
	
	We generate a nominal motion command by tracking $q_i^\star$ with a
	finite-horizon LQ controller applied to the double-integrator model of
	$p_i$. Define the state and input
	\[
	x_{i}=[p_{i,x},\dot p_{i,x},p_{i,y},\dot p_{i,y}]_k^\top,
	\qquad
	a_{i}=[a_{i,x},a_{i,y}]^\top,
	\]
	where $a_{i}$ is a fictitious Cartesian acceleration, not yet a physical
	input. Discretizing with sampling period $T$ gives
	\begin{equation}
		x_{i}^{k+1}=A_{i,d}x_{i}^{k}+B_{i,d}a_{i}^{k},
		\label{eq:double-int}
	\end{equation}
	with
$
\footnotesize
            A_{i,d}=\operatorname{blkdiag}(A_{i,s},A_{i,s}),\,  B_{i,d}=\begin{bmatrix}B_{i,s}&0\\0&B_{i,s}\end{bmatrix}$,\,
$			
\footnotesize
A_{i,s}=\begin{bmatrix}1&T\\0&1\end{bmatrix},
$
$		
\footnotesize
B_{i,s}=\begin{bmatrix}T^2/2\\T\end{bmatrix},
$
	where the symbol $\operatorname{blkdiag}(\cdot)$ constructs a block diagonal matrix from its matrix arguments. Equation~\eqref{eq:double-int} represents the standard zero-order-hold discretization of a double integrator, applied independently to the two spatial channels. The desired state is
	$
	\bar q_i^\star=
	[q_{i,x}^\star,0,q_{i,y}^\star,0]^\top,
	$
	which asks the agent to arrive at $q_i^\star$ with zero velocity. 
	
	Then, the objective function to minimize is constructed with a horizon $H$ as
	\begin{equation}
    \begin{aligned}
		J_i = \sum_{\ell =k}^{k+H-1} \left( (x_{i}^{\ell} - \bar q_i^\star)^\top Q_i (x_{i}^{\ell} - \bar q_i^\star) + (a_{i}^{\ell})^\top R_i a_{i}^{\ell} \right) \\+ (x_{i}^{k+H} - \bar q_i^\star)^\top Q_i (x_{i}^{k+H} - \bar q_i^\star),
		\label{eq:lqr-cost}
    \end{aligned}
	\end{equation}
	where 
	$
    \footnotesize
		Q_i= w_{\gamma_i}\gamma_i C_i^\top C_i+Q_{i,0},
		\;
		C_i=
		\begin{bmatrix}
			1&0&0&0\\
			0&0&1&0
		\end{bmatrix},
		\label{eq:lqr-weight}
	$
	The matrix $C_i$ extracts the position components of $x_i$, $w_{\gamma_i}>0$ is a
	tuning gain, and $Q_{i,0}\succeq0$ is a baseline weight that keeps the problem
	well-posed even when $\gamma_i=0$.
	
	The optimal finite-horizon control for tracking a fixed reference with
	quadratic cost is obtained by the usual backward Riccati recursion,
	carried through the value function $V_{i}^{k}(x_{i}^{k})=(x_{i}^{k})^\top P_{i}^{k}x_{i}^{k}+2(s_{i}^{k})^\top x_{i}^{k}+\text{const}$,
	where $P_{i}^{k} \succ 0$ is the symmetric positive-definite cost-to-go matrix, $s_{i}^{k}$ is the vector capturing the linear cross-term arising from the reference tracking offset, and `$\text{const}$' denotes a state-independent scalar term resulting from the constant offset in the tracking cost.
	For horizon $H$, this yields
	\begin{align}
		K_{i}^{k}&=(B_{i,d}^\top P_{i}^{k}B_{i,d}+R_i)^{-1},\notag\\
		P_{i}^{k-1}
		&=Q_i+A_{i,d}^\top
		(P_{i}^{k}-P_{i}^{k}B_{i,d}K_{i}^{k}B_{i,d}^\top P_{i}^{k})A_{i,d},\notag\\
		s_{i}^{k-1}
		&=-Q_i\bar q_i^\star+
		A_{i,d}^\top(s_{i}^{k}-P_{i}^{k}B_{i,d}K_{i}^{k}B_{i,d}^\top s_{i}^{k}),
		\label{eq:riccati}
	\end{align}
	initialized at the horizon end with $P_i^{H}=Q_i$ and
	$s_i^{H}=-Q_i\bar q_i^\star$, matching the terminal cost
	$(x_i^{H}-\bar q_i^\star)^\top Q_i(x_i^{H}-\bar q_i^\star)$. Carrying the
	recursion down to $P_i^1,s_i^1$ and applying the standard LQ minimizer to the
	current state $x_i$ gives the first control action,
	\begin{equation}
		a_i^{\rm nom}
		=-(B_{i,d}^\top P_{i}^{1}B_{i,d}+R_i)^{-1}
		(B_{i,d}^\top P_{i}^{1}A_{i,d}x_i+B_{i,d}^\top s_i^1).
		\label{eq:lqr-control}
	\end{equation}
	Rather than executing the full $H$-step open-loop plan, the problem is
	re-solved from scratch at every sampling instant using the current
	D$^2$OC target $q_i^\star$ and allocation weight $\gamma_i$, thus the
	controller behaves as a receding-horizon law that automatically adapts to
	the allocation changing over time. Note that since the weights of nearby 
	sample points $q_j$ decrease at every time step according to the D$^2$OC 
	update rule (refer to \cite{seo2025d2oc} for details), $q_i^\star$ dynamically evolves to drive the agent toward 
	uncovered regions, enabling closed-loop density-driven coverage.
	
\subsection{Physical Input Generation}

The nominal command $(a_{i}^{\rm nom})^{k}$ from \eqref{eq:lqr-control} is a
Cartesian acceleration command for the look-ahead point $p_i$, whereas the
unicycle is actuated by the physical inputs
$u_{i}^{k}=[v_{i}^{k},\omega_{i}^{k}]^\top$. The first-order
feedback-linearization relation \eqref{eq:fbl} specifies the velocity of
$p_i$ as a function of the physical input. To realize the Cartesian
acceleration command, we differentiate \eqref{eq:fbl} with respect to time,
yielding the continuous-time relation
\begin{equation}
\ddot p_i
=
G(\theta_i)
\begin{bmatrix}
\dot v_i\\
\dot\omega_i
\end{bmatrix}
+b_i,
\label{eq:fbl-second-order}
\end{equation}
where
\begin{equation}
b_i=
\begin{bmatrix}
-v_i\omega_i\sin\theta_i-d\omega_i^2\cos\theta_i\\
v_i\omega_i\cos\theta_i-d\omega_i^2\sin\theta_i
\end{bmatrix}.
\label{eq:fbl-drift}
\end{equation}
The term $b_i$ arises from the time variation of $G(\theta_i)$ and captures
the acceleration of the look-ahead point induced by the rotational motion of
the unicycle.

For the discrete-time implementation, let $t_k=kT$ and define
$u_{i}^{k}=u_i(t_k)$. Approximating the input-rate term over one sampling
interval by
\begin{equation}
\begin{bmatrix}
\dot v_i(t_k),
\dot\omega_i(t_k)
\end{bmatrix}^{\top}
\approx
\frac{u_{i}^{k}-u_{i}^{k-1}}{T},
\end{equation}
and evaluating the drift term using the previously applied input
$u_i^{k-1}$, define
\begin{equation}
\bar b_i^k
=
\begin{bmatrix}
-v_i^{k-1}\omega_i^{k-1}\sin\theta_i^k
-d(\omega_i^{k-1})^2\cos\theta_i^k\\
v_i^{k-1}\omega_i^{k-1}\cos\theta_i^k
-d(\omega_i^{k-1})^2\sin\theta_i^k
\end{bmatrix}.
\label{eq:fbl-drift-dt}
\end{equation}
Imposing the nominal acceleration condition
$\ddot p_i(t_k)=a_i^{\rm nom}$ then gives
\begin{equation}
\small
G(\theta_i^k)
\frac{(u_i^{\rm nom})^k-u_i^{k-1}}{T}
+\bar b_i^k
=
(a_i^{\rm nom})^k,
\label{eq:dt-fbl}
\end{equation}
and hence
\begin{equation}
(u_i^{\rm nom})^k
=
u_i^{k-1}
+
T G(\theta_i^k)^{-1}
\left(
(a_i^{\rm nom})^k-\bar b_i^k
\right).
\label{eq:inv-fbl}
\end{equation}
Thus, the Cartesian acceleration command generated by the discrete-time LQ
controller is converted into a nominal physical input using the previous
applied input and the feedback-linearization relation.

The nominal input $(u_i^{\rm nom})^k$ is not required to satisfy the actuator
constraints $u_i^k\in\mathcal U$ and does not account for neighboring agents.
Consequently, it serves only as the reference input for the physical-input
CBF-QP that will be explained later. The CBF-QP then determines the admissible physical
input $u_i^k\in\mathcal U$ while enforcing the collision-avoidance
constraints. Consequently, feedback linearization generates the nominal physical
input from information available before the current QP solution, and the
CBF-QP subsequently modifies that input directly in the physical control
space.

\section{Physical-Input Safety Filter}
\label{sec:safety-filter}

\subsection{Barrier Constraint}

For neighboring robots $i$ and $j$, define the look-ahead separation
$
\Delta p_{ij}(t)=p_i(t)-p_j(t)
$
and the barrier function
\begin{equation}
h_{ij}(t)=\|\Delta p_{ij}(t)\|^2-D_{\rm cbf}^2,
\label{eq:hij}
\end{equation}
so $h_{ij}(t)\ge0$ exactly when the two look-ahead points are at least
$D_{\rm cbf}$ apart. Differentiating \eqref{eq:hij} and substituting the
feedback-linearized dynamics \eqref{eq:fbl} for $\dot p_i(t)$, we have
\begin{equation}
\dot h_{ij}(t)
=
2\Delta p_{ij}(t)^\top G(\theta_i(t))u_i(t)
-2\Delta p_{ij}(t)^\top\dot p_j(t).
\label{eq:hij-dot}
\end{equation}
Writing $\Delta p_{ij}(t)=[\Delta x_{ij}(t),\Delta y_{ij}(t)]^\top$ and
expanding the matrix product $\Delta p_{ij}^\top G(\theta_i)$ componentwise,
\begin{equation}
\begin{split}
\Delta p_{ij}^\top G(\theta_i)
&=
\big[
\underbrace{\Delta x_{ij}\cos\theta_i+\Delta y_{ij}\sin\theta_i}_{\xi_{ij}},
\\
&\quad
\underbrace{d(-\Delta x_{ij}\sin\theta_i+\Delta y_{ij}\cos\theta_i)}_{\zeta_{ij}}
\big],
\end{split}
\label{eq:barrier-coeff}
\end{equation}
where all quantities are evaluated at the same time $t$. Thus,
\eqref{eq:hij-dot} becomes an expression that is affine in the physical
input $(v_i(t),\omega_i(t))$:
\begin{equation}
\dot h_{ij}(t)
=
2\xi_{ij}(t)v_i(t)
+2\zeta_{ij}(t)\omega_i(t)
-2\Delta p_{ij}(t)^\top\dot p_j(t).
\label{eq:hij-expanded}
\end{equation}
This affine structure is what allows the collision-avoidance requirement
to be enforced as a linear constraint inside a convex QP rather than a
nonlinear one. Requiring the exponential CBF condition
$\dot h_{ij}(t_k)\ge-\alpha h_{ij}(t_k)$, with $\alpha>0$ a design gain,
at each sampling instant $t_k=kT$, agent $i$ solves
\begin{equation}
\begin{aligned}
(u_{i}^\star)^{k}
=\arg\min_{u_i\in\mathcal U}\;&
\|u_i-u_i^{\rm nom}(t_k)\|^2\\
\text{s.t.}\;
2\xi_{ij}(t_k)v_i+2\zeta_{ij}(t_k)\omega_i
&\ge
2\Delta p_{ij}(t_k)^\top\dot p_j(t_k)
-\alpha h_{ij}(t_k),
\\[-1mm]
\hspace{12mm}
\forall j:\ &\|p_i(t_k)-p_j(t_k)\|<R_s,
\end{aligned}
\label{eq:cbfqp}
\end{equation}
where $R_s$ denotes the interaction (or sensing) range within which neighboring agents are considered for collision avoidance.

The same decision variable $u_i$ is thus simultaneously the commanded
physical input and the input appearing in the safety certificate, with no
intermediate re-mapping step that could introduce conservatism or
infeasibility mismatches. When no neighbor is within $R_s$, all
constraints vanish and \eqref{eq:cbfqp} reduces to the Euclidean
projection of the nominal input onto the box $\mathcal U$.

\subsection{Input Saturation and Avoidance Authority}

Because \eqref{eq:cbfqp} is a constrained QP over the bounded box
$\mathcal U$ (where $v_i \in [0, v_{i,\max}]$ and
$\omega_i \in [-\omega_{i,\max}, \omega_{i,\max}]$), its feasibility depends
on the support function of $\mathcal U$ in the barrier normal direction
$(\xi_{ij},\zeta_{ij})$. Specifically, since the linear velocity is
restricted to a unilateral interval $[0,v_{i,\max}]$, its optimal
contribution is active only when $\xi_{ij}>0$, yielding
$2\max(\xi_{ij},0)v_{i,\max}$. Conversely, since the angular velocity is
bounded symmetrically in $[-\omega_{i,\max},\omega_{i,\max}]$, its
contribution is maximized by aligning its sign with $\zeta_{ij}$, resulting
in $2|\zeta_{ij}|\omega_{i,\max}$. Thus, the maximum value the left-hand
side of the constraint in \eqref{eq:cbfqp} can attain is
\begin{equation}
\max_{u_i\in\mathcal U}
(2\xi_{ij}v_i+2\zeta_{ij}\omega_i)
=
2\max(\xi_{ij},0)v_{i,\max}
+2|\zeta_{ij}|\omega_{i,\max}.
\label{eq:support}
\end{equation}

In a head-on encounter, the neighbor lies ahead along the current heading,
so that $\xi_{ij}<0$ and the forward-velocity channel cannot increase the
barrier value at that instant. The available avoidance authority for the
look-ahead-point barrier therefore comes entirely from the angular-rate
channel. Since
$|\zeta_{ij}|\le d\|\Delta p_{ij}\|$ by \eqref{eq:barrier-coeff}, its
maximum contribution is bounded by
$
2d\|\Delta p_{ij}\|\omega_{i,\max}.
$
Thus, the look-ahead construction provides a nonzero angular avoidance
channel even when the unilateral forward-velocity constraint prevents the
linear-velocity channel from contributing. Whether this authority is
sufficient for a particular configuration depends on the barrier geometry,
the current separation, and the remaining terms in the CBF constraint.

This single-neighbor characterization does not automatically extend to the
multi-neighbor case, where the intersection of multiple CBF half-spaces
with $\mathcal U$ may be empty even if each constraint is individually
feasible. This is the multi-neighbor feasibility issue studied in
\cite{zhang2024feasibility}. In our implementation, an infeasible QP causes
the robot to stop ($v_i=0,\omega_i=0$). This respects the actuator limits,
but does not by itself guarantee
$\dot h_{ij}(t_k)\ge-\alpha h_{ij}(t_k)$ and is therefore only a safe
degradation mechanism.

The above analysis characterizes the control authority and feasibility of
the CBF constraint in the look-ahead-point space. Unfortunately, it does not by
itself guarantee physical collision avoidance, since the barrier is defined
on $p_i$ while the physical robot bodies are located relative to $r_i$.
The resulting physical clearance is addressed separately in the following
subsection.

\subsection{Physical Clearance and Discrete-Time Margin}
\label{sec:margin}

The barrier \eqref{eq:hij} is defined on the look-ahead points $p_i$, whereas
the physical robot bodies are located relative to the rear-axle positions
$r_i$. This subsection therefore translates the look-ahead-point separation
enforced by the CBF into a conservative physical-clearance condition.
The resulting margin accounts for both the geometric offset introduced by
the look-ahead construction and the motion that can occur between
consecutive control updates.

\paragraph{Geometric offset}
From \eqref{eq:lookahead}, $r_i=p_i-d\,e_i$ with
$e_i=[\cos\theta_i,\sin\theta_i]^\top$, thus
$
r_i-r_j=(p_i-p_j)-d(e_i-e_j).
$

Since $\|e_i\|=\|e_j\|=1$, the triangle inequality gives
$\|e_i-e_j\|\le2$, and therefore
\begin{equation}
\norm{r_i-r_j}
\ge
\norm{p_i-p_j}-2d.
\label{eq:geometric-margin}
\end{equation}
This bound converts a guaranteed separation of the look-ahead points into
a conservative lower bound on the separation of the rear axles, with a
worst-case loss of $2d$ due to the relative headings.

\paragraph{Discrete-time motion}
From \eqref{eq:fbl},
$G(\theta)=R(\theta)\operatorname{diag}(1,d)$, where $R(\theta)$ is an
orthogonal rotation matrix. Therefore,
$
\norm{\dot p_i}
=
\norm{\operatorname{diag}(1,d)u_i}
=
\sqrt{v_i^2+d^2\omega_i^2},
$
and under the actuator box, this speed is bounded by
\begin{equation}
\norm{\dot p_i}\le
V_p:=
\sqrt{v_{\max}^2+d^2\omega_{\max}^2}.
\label{eq:speed-bound}
\end{equation}
Let $t_k=kT$ denote the sampling instants. Under zero-order hold over one
sampling interval of length $T$, the look-ahead point can move by at most
$V_pT$ between consecutive samples:
$
\norm{p_i(t)-p_i(t_k)}\le V_pT,
\qquad
t\in[t_k,t_{k+1}].
$
Applying the reverse triangle inequality and then the triangle inequality,
we obtain
\begin{equation}
\begin{aligned}
&\left|
\norm{p_i(t)-p_j(t)}
-
\norm{p_i(t_k)-p_j(t_k)}
\right|\\
% &\le
% \norm{
% \bigl(p_i(t)-p_i(t_k)\bigr)
% -
% \bigl(p_j(t)-p_j(t_k)\bigr)
% }\\
&\le
\norm{p_i(t)-p_i(t_k)}
+
\norm{p_j(t)-p_j(t_k)}
\le
2V_pT
=:\delta.
\end{aligned}
\label{eq:sample-margin}
\end{equation}

\paragraph{Combined design condition}
Suppose each robot's physical body is contained in a disk of radius $\rho$
centered at $r_i$, and let $C_{\rm req}$ denote the desired body-to-body
clearance. Then
$D_s:=2\rho+C_{\rm req}$ is the minimum acceptable distance between the
centers $r_i$ and $r_j$. If the CBF-QP remains feasible and enforces
\[
\norm{p_i(t_k)-p_j(t_k)}\ge D_{\rm cbf}
\]
at every sampling instant, then \eqref{eq:sample-margin} implies that, for
all $t\in[t_k,t_{k+1}]$,
$
\norm{p_i(t)-p_j(t)}
\ge
D_{\rm cbf}-2V_pT.
$

Combining this bound with \eqref{eq:geometric-margin} yields
$
\norm{r_i(t)-r_j(t)}
\ge
D_{\rm cbf}-2V_pT-2d.
$
Therefore, a sufficient design condition for the desired physical clearance
is
\begin{equation}
D_{\rm cbf}
\ge
D_s+2d+2V_pT.
\label{eq:margin}
\end{equation}
Thus, the required CBF margin consists of three contributions: the
look-ahead geometry $2d$, the discrete-time motion $2V_pT$, and the desired
physical clearance $D_s$.

The condition \eqref{eq:margin} is a geometric discrete-time margin and
does not replace the continuous-time CBF condition. The CBF constraint
\eqref{eq:cbfqp} is derived from the continuous-time condition
$\dot h_{ij}\ge-\alpha h_{ij}$ and is evaluated at each sampling instant.
Equation \eqref{eq:margin} additionally accounts for the maximum separation
that can be lost between two consecutive updates under the actuator bounds.
Hence, when the CBF-QP remains feasible and maintains
$\norm{p_i(t_k)-p_j(t_k)}\ge D_{\rm cbf}$ at the sampling instants, the
margin in \eqref{eq:margin} provides a conservative physical-clearance
guarantee between updates by accounting for both the look-ahead geometry and
the bounded discrete-time motion.

	\section{Integrated Algorithm}
	
The complete per-agent procedure is summarized in
Algorithm~\ref{alg:main}. The D$^2$OC allocation, nominal LQ controller,
feedback-linearization step, and CBF-QP are executed at every sampling
instant. Only neighbor states within $R_s$ are required by the safety
filter, while the D$^2$OC vectors $\gamma_i$ used for task allocation are synchronized when agents communicate (if within the communication range $R_c$).

	\begin{algorithm}[!h]
    \small
		\caption{Nonholonomic D$^2$OC with Physical-Input CBF Filter}
		\label{alg:main}
		\begin{algorithmic}[1]
			\State Compute $p_i=r_i+d[\cos\theta_i,\sin\theta_i]^\top$ and
			$\dot p_i=G(\theta_i)u_i$
			\State Update local D$^2$OC weights and compute $q_i^\star,\gamma_i$
			using \eqref{eq:d2oc-target}
			\State Solve the finite-horizon LQ problem and obtain
			$a_i^{\rm nom}$ from \eqref{eq:lqr-control}
			\State Convert $a_i^{\rm nom}$ to $u_i^{\rm nom}$ using
			\eqref{eq:inv-fbl}
			\State Gather $(p_j,\dot p_j)$ for neighbors within $R_s$
			\State Solve the CBF-QP \eqref{eq:cbfqp}
			\State If infeasible, set $u_i^\star=(0,0)^\top$
			\State Apply $u_i^\star$ and propagate \eqref{eq:unicycle}
			\State Synchronize D$^2$OC weights with communicating neighbors
		\end{algorithmic}
	\end{algorithm}
	
	\section{Simulation and Experimental Results}

To validate the efficacy and real-world applicability of the proposed density-driven optimal control framework, we conduct both numerical simulations and hardware experiments using Robotarium platforms. 
We evaluate a system of unicycle robots covering a target density represented by an eight-component Gaussian mixture, where the Gaussian means are sampled uniformly from $[20,80]^2$ with covariance $10I$, and the resulting target distribution is discretized using $N=4000$ sample points. 
For the numerical simulations, initial robot positions are independently sampled from $[10,90]^2$ with random headings, running for 5000 control steps across five independent random seeds (0--4). 
For the experimental validation, we deploy the control algorithms onto physical multi-robot hardware to demonstrate robust obstacle avoidance and trajectory tracking under realistic actuator constraints. Both simulation and robotarium parameters are shown in Table \ref{tab:params}.
	
	\begin{table}[!h]
		\centering
		\caption{Controller parameters for the simulation study and for the
			Robotarium hardware deployment.}
		\label{tab:params}
		\footnotesize
		\begin{tabular}{@{}l@{\hspace{6pt}}l@{\hspace{4pt}}c@{\hspace{6pt}}c@{}}
			\toprule
			Parameter & Description & Simulation & Robotarium\\
			\midrule
			$n_a$ & number of robots & 5 & 5\\
			$T$ & sampling time [s] & 0.1 & 0.066\\
			$R_s$ & interaction range [m] & 15 & 0.30\\
            $R_c$ & communication range [m] & 15 & -\\
			$d$ & look-ahead offset [m] & 0.4 & 0.008\\
			$D_s$ & required separation [m] & 2.5 & 0.200\\
			$D_{\rm cbf}$ & CBF design distance [m] & 4.11 & 0.227\\
			$V_p$ & maximum look-ahead speed [m/s] & 4.05 & 0.081\\
			% $\delta$ & sampled-data margin [m] & 0.81 & 0.011\\
			$\alpha$ & CBF gain & 2.0 & 2.0\\
			$H$ & LQ horizon & 10 & 10\\
			$Q_0$ & LQ regularizer & $0.02I_4$ & $0.02I_4$\\
			$w_{\gamma_i}$ & target-weight gain & 3000 & 3000\\
			$R$ & control weight & $1.5I_2$ & $1.5I_2$\\
			$v_{\min},v_{\max}$ & linear speed [m/s] & $[0,4.0]$ & $[0,0.08]$\\
			$\omega_{\max}$ & angular speed [rad/s] & $\pi/2$ & $\pi/2$\\
			\bottomrule
		\end{tabular}
	\end{table}
	
\subsection{Two-Layer Baseline}

To isolate the effect of enforcing safety directly on the physical input,
we compare the proposed method with a conventional two-layer architecture
using the same D$^2$OC allocation and finite-horizon LQ nominal controller.
The baseline uses the same look-ahead point $p_i$ as the virtual motion
variable. Then, a higher-order control barrier function quadratic program (HOCBF-QP) \cite{xiao2019hocbf} generates a collision-free virtual
trajectory by enforcing
\begin{equation}
\ddot h_{ij}+2\alpha\dot h_{ij}+\alpha^2h_{ij}\ge0,
\qquad
h_{ij}=\|p_i-p_j\|^2-D_{\rm cbf}^2 .
\label{eq:baseline_hocbf}
\end{equation}
The same $D_{\rm cbf}$ is used for both methods so that the comparison
does not change the prescribed virtual safety distance.

The resulting virtual look-ahead trajectory is then passed to a separate
low-level tracking controller. Let $(p_i^{\rm vir})^k$ and
$(v_{p,i}^{\rm vir})^k$ denote the virtual look-ahead position and Cartesian
velocity at sampling instant $t_k=kT$. The physical input is computed as
$
(u_i^{\rm track})^k
=
G(\theta_i^k)^{-1}
\left[
(v_{p,i}^{\rm vir})^k
+
K_p\left((p_i^{\rm vir})^k-p_i^k\right)
\right],
\label{eq:baseline}
$
and independently clipped to $\mathcal U$, where $K_p=3$ is used in our simulation.
Thus, the HOCBF constrains the virtual trajectory, but does not constrain
the physical input that is ultimately applied to the robot.

This is referred to as a \emph{two-layer} architecture because safety is
handled in a virtual planning layer and trajectory realization is handled
separately by a low-level tracking layer. In contrast, the proposed method
maps the nominal LQ command to the physical input and applies the CBF
constraint directly to $(v_i,\omega_i)$ in the same QP that enforces
$\mathcal U$. No separate safety-certified virtual trajectory is introduced
between the nominal controller and the physical input, which makes the
proposed architecture \emph{single-layer}.

	\subsection{Simulation Results}

\begin{figure}[t]
    \subfloat[]{
        \includegraphics[width=0.495\linewidth]{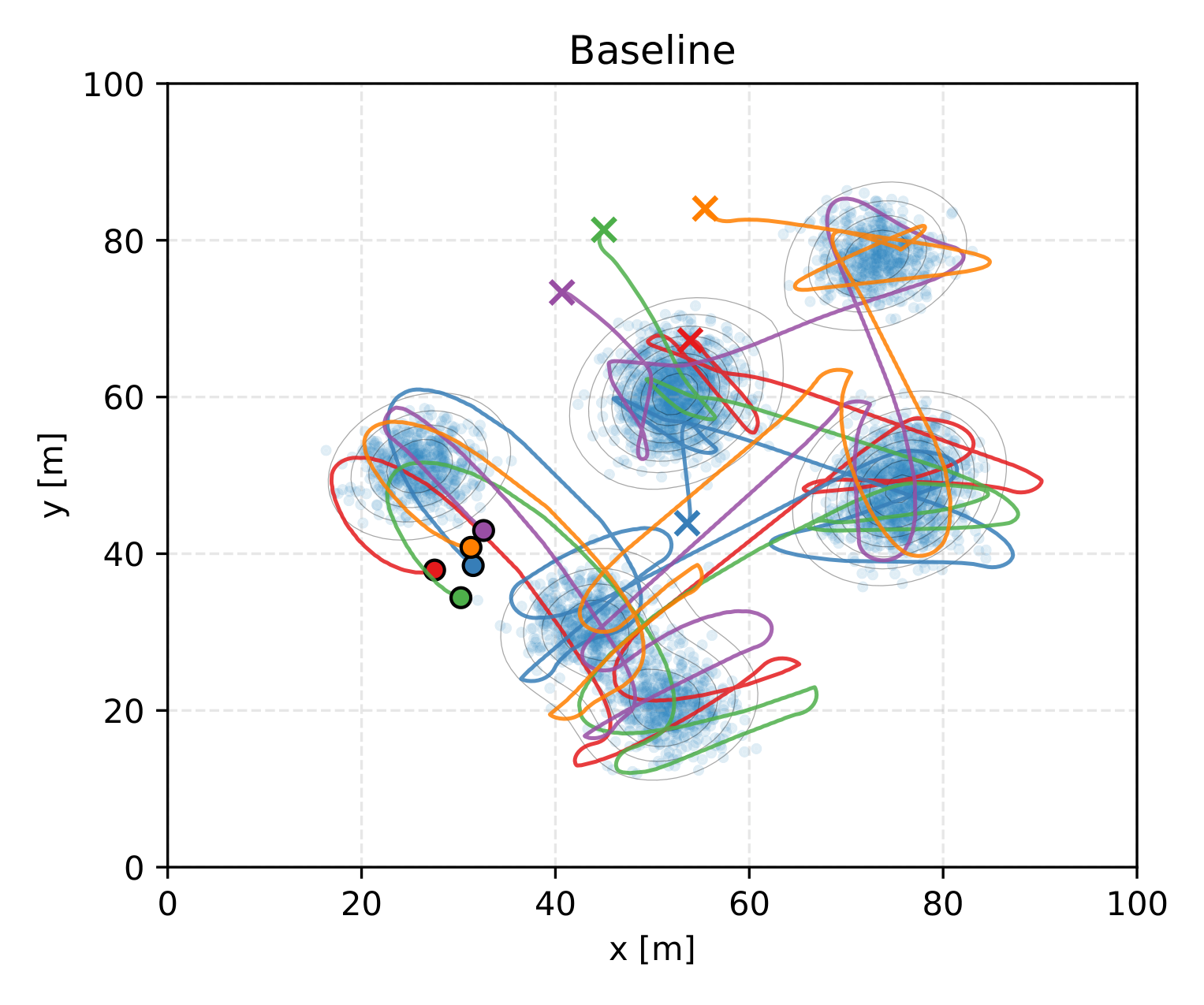}}
    \subfloat[]{
        \includegraphics[width=0.495\linewidth]{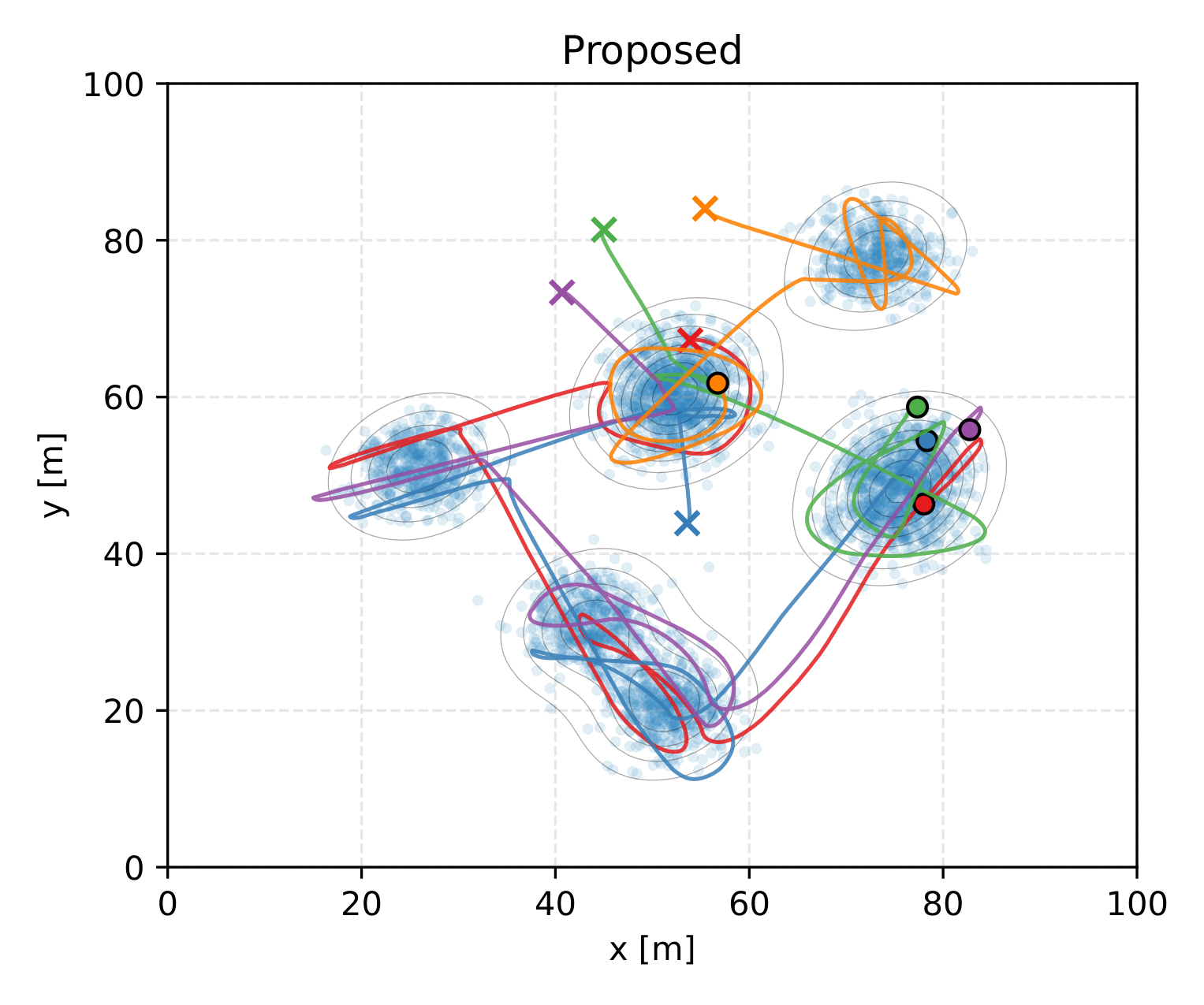}}
    \caption{Five-robot trajectories over the reference density for (a) the
    baseline and (b) the proposed method; Legend: $\times$ and $\circ$ denote initial
    and final positions, respectively.}
    \label{fig:snapshot1}
\end{figure}

Fig.~\ref{fig:snapshot1} presents the robot trajectories over the reference
density. Both methods move toward under-covered high-density regions and
perform short avoidance maneuvers when robots approach one another, with
similar overall coverage behavior.

\begin{figure}[!t]
    \centering
    \includegraphics[width=\linewidth]{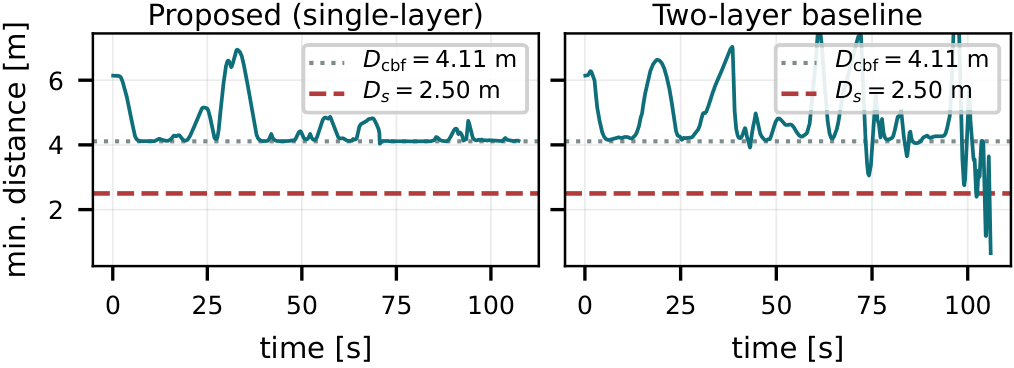}
    \caption{Minimum inter-robot distance. The proposed method maintains the
    look-ahead separation near $D_{\rm cbf}=4.11$ m, whereas the two-layer
    baseline falls below the required physical distance $D_s=2.50$ m.}
    \label{fig:mindist}
\end{figure}

Fig.~\ref{fig:mindist} compares the minimum inter-robot distance at the
look-ahead points. In this plot, the proposed method maintains
the look-ahead distance near $D_{\rm cbf}=4.11$ m, with a corresponding
rear-axle minimum of $3.85$ m. The two-layer baseline instead reaches
$0.66$ m and $1.27$ m, respectively, showing that its virtual safety
certificate does not transfer to the physical robot under tracking error
and actuator saturation.

Table~\ref{tab:comparison} summarizes the quantitative results over five different trials. The proposed
method never violates $D_s$, achieving a minimum physical distance of
$3.89\pm0.08$ m, while the baseline reaches $0.96\pm0.75$ m and violates
$D_s$ in all five trials. Coverage and completion time remain comparable,
whereas the proposed method produces fewer QP infeasibility events. 

\begin{table}[h]
    \centering
    \caption{Safety comparison over five trials. Mean $\pm$ standard deviation is reported for all repeated quantities. ``Worst'' denotes the minimum value observed across the five trials.}
    \label{tab:comparison}
    \footnotesize
    \begin{tabular}{@{}lcc@{}}
        \toprule
        & Proposed & Two-layer\\
        \midrule
        % Final coverage [\%] & $99.92\pm0.01$ & $99.93\pm0.01$\\
        Completion time [s] & $109.6\pm5.0$ & $106.0\pm3.3$\\
        Min. physical distance [m] & $3.89\pm0.08$ & $0.96\pm0.75$\\
        Worst physical distance [m] & $3.76$ & $0.075$\\
        Steps below $D_s$ & $0.0\pm0.0\%$ & $3.5\pm3.3\%$\\
        Trials with violation & $0/5$ & $5/5$\\
        QP infeasibility count & $12.4\pm6.2$ & $53.4\pm30.3$\\
        \bottomrule
    \end{tabular}
\end{table}

\begin{figure}[t]
		\centering
		\includegraphics[width=.9\linewidth]{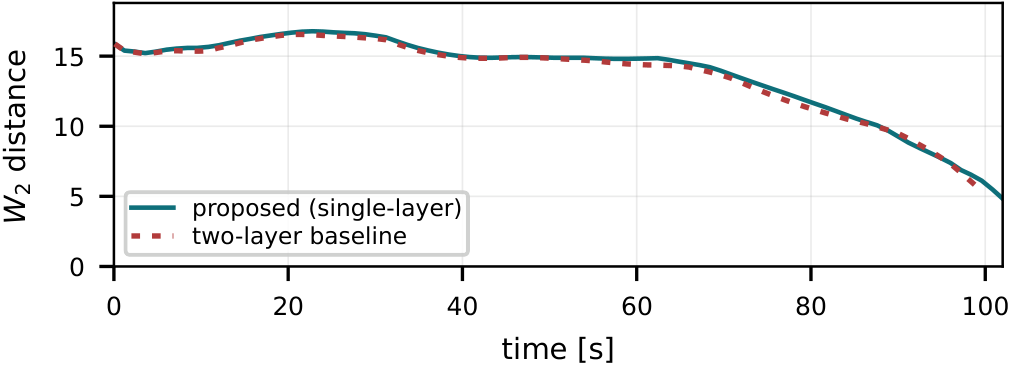}
        \caption{$2$-Wasserstein distance to the reference density:
		proposed method (solid) and two-layer baseline (dotted).}
		\label{fig:sim_w2}
	\end{figure}

To quantitatively compare the coverage performance of the two methods, the $2$-Wasserstein
	distance between the empirical distribution formed by the multi-agent time-averaged trajectories and the reference density is
	computed. A smaller Wasserstein distance indicates better alignment between the empirical distribution and the reference density. 
	  Fig.~\ref{fig:sim_w2} shows it falling from $15.9$ to $4.8$ for the
	proposed method and to $5.4$ for the baseline in a non-monotonic fashion. 
	Although a noticeable trajectory difference is observed in Fig.~\ref{fig:snapshot1}, the Wasserstein distance plot in Fig.~\ref{fig:sim_w2} presents minimal difference. 
	Despite this similarity in distribution matching, the baseline fails to maintain 
	safety and experiences multiple collisions, whereas the proposed method guarantees collision-free navigation as guaranteed in Fig.~2 and Table~II.
	
	\subsection{Hardware Validation}
	\label{sec:hardware}
	
	To assess whether the single-layer architecture transfers to physical
	hardware, the controller was deployed on the Robotarium
	\cite{pickem2017robotarium}, a remotely accessible multi-robot testbed, using
	five differential-drive robots in a $3.2\times2.0$ m arena. 
    Note that the Robotarium requires controllers to pass its simulation-based safety validation before hardware deployment. Since the two-layer baseline fails
this validation, as observed in the simulation result, it could not be deployed to the Robotarium. Thus, the hardware experiment evaluates the proposed controller only.
    The D$^2$OC
	allocation, the finite-horizon LQ tracker, and the CBF-QP are unchanged from
	the simulation study, and the LQ weights $Q_0$, $R$, $w_{\gamma_i}$, and the
	horizon $H$ are identical. Only the kinematic and geometric constants are
	re-instantiated at hardware scale, as listed in Table~\ref{tab:params}. Rather than rescaling the simulation
	value, $D_s$ is obtained from the physical robot, with body radius
	$\rho=0.055$ m and required body-to-body clearance $C_{\rm req}=0.09$ m,
	giving $D_s=2\rho+C_{\rm req}=0.200$ m. 
	The Robotarium actuator limits and sampling
	period give $V_p=\sqrt{v_{\max}^2+d^2\omega_{\max}^2}=0.081$ m/s and a
	sampled-data term $2V_pT=0.011$ m. In turn, with $D_s=0.200$ m and $2d=0.016$ m,
	condition \eqref{eq:margin} requires $D_{\rm cbf}\ge0.227$ m. 
	
	The Robotarium arena is bounded, whereas the simulation domain is not. The
	CBF-QP is therefore augmented with four static half-space constraints, one
	per wall, expressed at the same look-ahead point $p_i$ and using the same
	class-$\mathcal{K}$ form as the inter-robot barrier in
	\eqref{eq:hij-dot}. Because the walls are stationary, these constraints
	carry no drift term. A boundary margin of $0.12$ m and a gain of $2.0$ were
	used. The wall constraints use a boundary margin of $0.12$ m and the same
class-$\mathcal{K}$ CBF form as the inter-robot constraint, with gain
$\alpha_w=2.0$. They enter the same QP as the collision and actuator
constraints, so the filtered command remains the solution of a single
optimization over the physical inputs $(v,\omega)$.

\begin{figure}[t]
		\centering
		\includegraphics[width=0.8\linewidth]{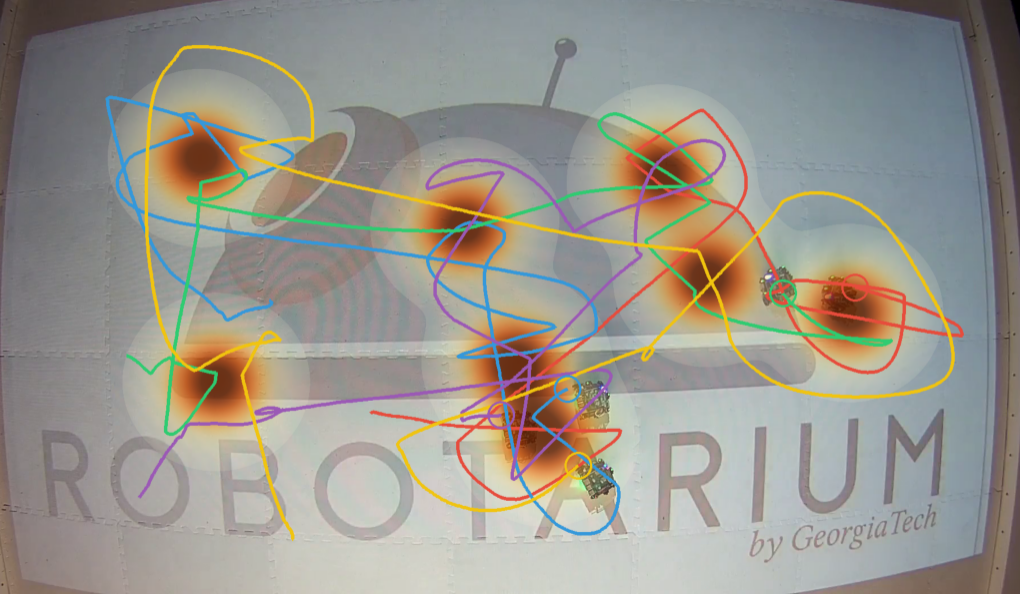}
        \caption{Robotarium deployment at the end of the run. The reference density
	   and the completed robot paths are overlaid onto the overhead camera view,
	   with one color per robot.}
		\label{fig:hw_overlay}
	\end{figure}

Fig.~\ref{fig:hw_overlay} shows an overlay of the reference density and the
five-agent trajectories on the Robotarium overhead camera view at the end of
the experiment. The agents exhibit sustained motion patterns consistent
with the spatial distribution prescribed by the reference density, with the
resulting time-averaged trajectories matching the high-density regions.
Coverage is completed after $4654$ control steps, corresponding to $307.2$ s.
	
\begin{figure}[t]
		\centering
		\includegraphics[scale=0.75]{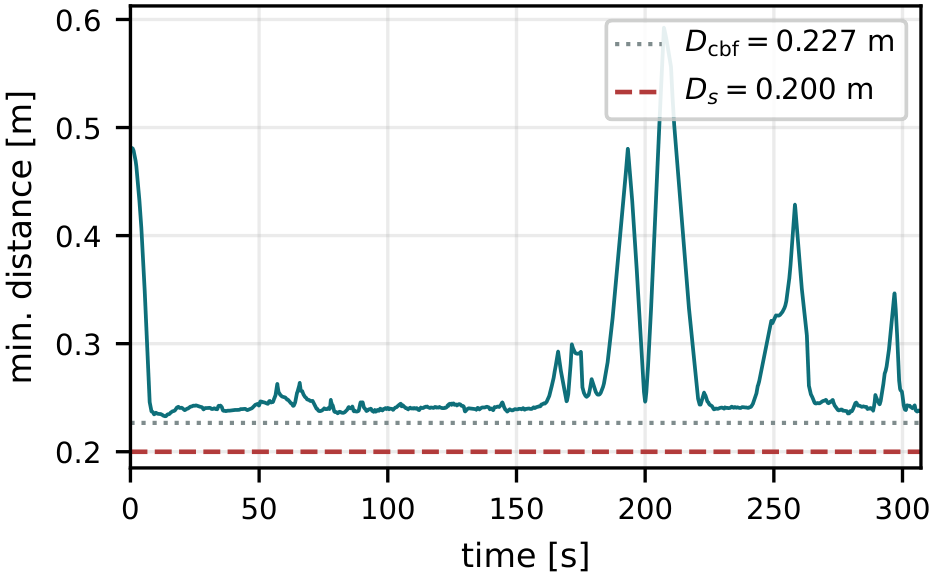}
		\caption{Minimum inter-robot distance on hardware, evaluated at the look-ahead
			points $p_i$ on which the barrier acts. The recorded minimum of $0.2369$ m
			stays above both $D_s=0.200$ m and $D_{\rm cbf}=0.227$ m for the entire
			trial; the corresponding rear-axle minimum is $0.2327$ m.}
		\label{fig:hw_mindist}
	\end{figure}

In addition, Fig.~\ref{fig:hw_mindist} shows the minimum inter-robot distance, following the convention of Fig.~\ref{fig:mindist}.
The separation remains above $D_{\rm cbf}$ throughout the trial, reaching a
minimum of $0.2369$ m. The corresponding  minimum is $0.2327$ m,
while the minimum over five deployments is $0.2312\pm0.0026$ m. No sample
falls below $D_s$, and no safety-monitor intervention occurs.

\begin{figure}[t]
		\centering
		\includegraphics[scale=0.8]{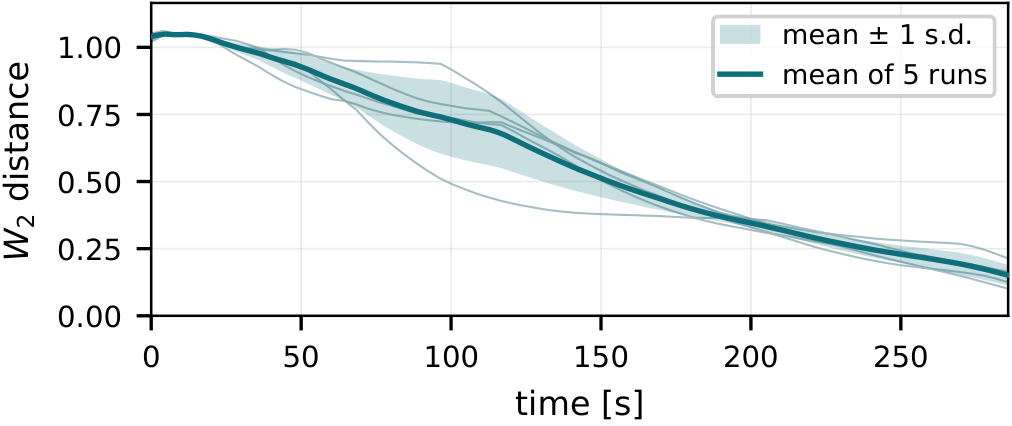}
        \caption{$2$-Wasserstein distance between the time-averaged spatial
        distribution induced by the five-agent trajectories and the reference
        density. Mean over five Robotarium deployments from identical initial poses,
        with a band of one standard deviation. The mean decreases from $1.04$ to
        $0.15$.}
		\label{fig:hw_w2}
	\end{figure}

For the quantitative measure, Fig.~\ref{fig:hw_w2} shows the $2$-Wasserstein distance between the
time-averaged spatial distribution induced by the five-agent trajectories
and the reference density over five independent runs from the same initial pose.
The mean decreases from $1.04$ to $0.15$, with the standard deviation
remaining below $0.14$, indicating consistent convergence of the
time-averaged distribution toward the reference density.

	\subsection{Limitations and Design Implications}

The experiments highlight three observations. First, the two-layer baseline fails because its CBF certifies a virtual
trajectory that the physical robot cannot always track under actuator limits. The proposed single-layer controller instead enforces the CBF
directly on $(v_i,\omega_i)$, ensuring that the executed motion itself satisfies the safety constraint.

Second, the proposed safety distance is not selected solely by empirical
tuning. The value used on the Robotarium accounts for both the look-ahead
offset and the maximum motion possible during one sampling interval, whereas
a margin based only on the geometric offset does not account for motion
between control updates.

Third, the current guarantee still depends on feasibility of the physical
CBF-QP. Simultaneous constraints from multiple neighbors can make the
intersection of the safety constraints and actuator limits empty.
Moreover, when two look-ahead points become close, the robot cannot reverse
and must rely on turning to increase separation. When multiple neighbors become close, the feasible set of the physical
CBF-QP can become empty because the available turning authority is limited
by the platform's actuation constraints. Thus, recovery is fundamentally
limited by physical control authority rather than by the barrier gain alone.

	\section{Conclusion}

This paper presented a single-layer density-driven coverage and safety architecture
for nonholonomic multi-robot systems. The proposed framework combines
D$^2$OC-based density allocation with finite-horizon LQ tracking and a
CBF-QP that directly filters the physical unicycle inputs $(v,\omega)$ under
their actuator limits. By enforcing safety at the physical input level, the
method avoids the gap between a safety-certified virtual trajectory and the
motion actually executed by the robot.

A look-ahead framework was formulated to connect the physical clearance
requirement to the CBF constraint through an explicit margin that accounts
for the robot geometry and its motion over each sampling interval. This
provides a direct way to select the CBF safety distance while retaining the
single-layer control structure. Simulation results show that the proposed
method preserves physical separation without sacrificing coverage
performance, whereas the two-layer baseline can lose physical safety when
its tracking layer cannot realize the certified virtual motion. Experiments
on the Robotarium further demonstrate that the proposed controller maintains
the required physical clearance during density-driven coverage with real
actuation and sensing.

Future work will focus on establishing formal feasibility conditions for
multiple simultaneous safety constraints and strengthening the safety
analysis under time discretization.

{\footnotesize
\bibliographystyle{IEEEtran}
\bibliography{reference}
}
	
\end{document}